\documentclass[11pt,a4paper]{article}
\usepackage[utf8]{inputenc} 
\usepackage[T1]{fontenc}    
\usepackage{hyperref}       
\usepackage{url}            
\usepackage{amsfonts}       
\usepackage{nicefrac}       
\usepackage{microtype}      
\usepackage{cleveref}       
\usepackage{lipsum}         
\usepackage{graphicx}
\usepackage{doi}
\usepackage{authblk}

\begin{document}

\title{A direct time-of-flight image sensor with in-pixel surface detection and dynamic vision}

\author[1,4,*]{Istvan Gyongy}
\author[1,4]{Ahmet T. Erdogan}
\author[2]{Neale A.W. Dutton}
\author[1]{Germán Mora Martín}
\author[1]{Alistair Gorman}
\author[1]{Hanning Mai}
\author[1,3]{Francesco Mattioli Della Rocca}
\author[1]{Robert K. Henderson}

\affil[1]{\small{School of Engineering, Institute for Integrated Micro and Nano Systems, The University of Edinburgh, Edinburgh, EH9 3FF, UK}}
\affil[2]{STMicroelectronics Imaging Division, Tanfield, Edinburgh, EH3 5DA, UK}
\affil[3]{Now with Sony Europe Technology Development Centre, 38123 Trento, Italy}
\affil[4]{The authors contributed equally to this work.}
\affil[*]{\url{istvan.gyongy@ed.ac.uk}}
\date{}



\maketitle

\begin{abstract}
3D flash LIDAR is an alternative to the traditional scanning LIDAR systems, promising precise depth imaging in a compact form factor, and free of moving parts, for applications such as self-driving cars, robotics and augmented reality (AR). Typically implemented using single-photon, direct time-of-flight (dToF) receivers in image sensor format, the operation of the devices can be hindered by the large number of photon events needing to be processed and compressed in outdoor scenarios, limiting frame rates and scalability to larger arrays. We here present a $64 \times 32$ pixel ($256 \times 128$ SPAD) dToF imager that overcomes these limitations by using pixels with embedded histogramming, which lock onto and track the return signal. This reduces the size of output data frames considerably, enabling maximum frame rates in the 10~kFPS range or 100~kFPS for direct depth readings. The sensor offers selective readout of pixels detecting surfaces, or those sensing motion, leading to reduced power consumption and off-chip processing requirements. We demonstrate the application of the sensor in mid-range LIDAR. 
\end{abstract}

\section{Introduction}

Direct time-of-flight (dToF) sensors measure depth by illuminating the scene of interest with a pulsed laser source, and timing the back-scattered return signal. The sensors are capable of centimetre precision over hundreds of meters and beyond, and as such are a key technology for LIDAR-based 3D vision systems providing situational awareness in self-driving cars and other autonomous systems \cite{rapp2020,roriz2021automotive}. 

With the use of single-photon avalanche diodes (SPADs) in complementary metal-oxide semiconductor (CMOS) technology, detector arrays with integrated processing can be realised, for a single-chip, all-digital receiver unit. Combining a SPAD receiver, implemented in image sensor format, with flood illumination provides a solid-state 3D imaging solution, with increased robustness to vibrations and reduced calibration needs compared with approaches involving mechanical scanning \cite{wang2020}. Furthermore, by capturing the entire field of view simultaneously, motion artefacts resulting from dynamic scenes are minimised. 

The premise of SPADs in 3D flash LIDAR has lead to significant research efforts of late, aiming to overcome the key challenges in developing viable, array format sensors for mid- to long range outdoor applications \cite{zhang2018, hutchings2019,kim2021,stoppa2021,patanwala2021,seo2021,kumagai2021,padmanabhan2021,zhang2021} . These challenges include equipping the sensor with a sufficient photon throughput to accommodate the maximal data rate of several tens of millions of events per second that can be generated by each SPAD. A second, related problem is ensuring an appropriate level of on-chip data compression to overcome readout bottleneck issues (as well as potential high memory requirements) and enable depth imaging at video rates and beyond. The overall sensor architecture should ideally be scalable to large array sizes without incurring excessive power consumption. 

Recent reviews of SPAD architectures for dToF/LIDAR include \cite{Villa2021,Piron2021,Gyongy2022}. In view of the high photon throughput and data compression requirements, one of the key trends has been embedded processing in the form of histogramming of the time-of-arrival of photons, with some architectures achieving higher still compression by processing these histograms to compute depth maps on-chip. The histogram processing can be carried out in-pixel, or outside the pixel array. In the latter case, there is more area available for histogram memory, but transferring data out of the pixel array can become challenging for large arrays. In either case, the accumulation of full histograms, covering the entire distance range of interest, can be impractical, not just because of the large memory requirements, but due to the increased power consumption and potential bottlenecks associated with the transfer of large amounts of data in and out of memory.

\begin{figure}[htpb]
\centering
\includegraphics[trim=0cm 2cm 0cm 2cm, clip=true, width=\linewidth]{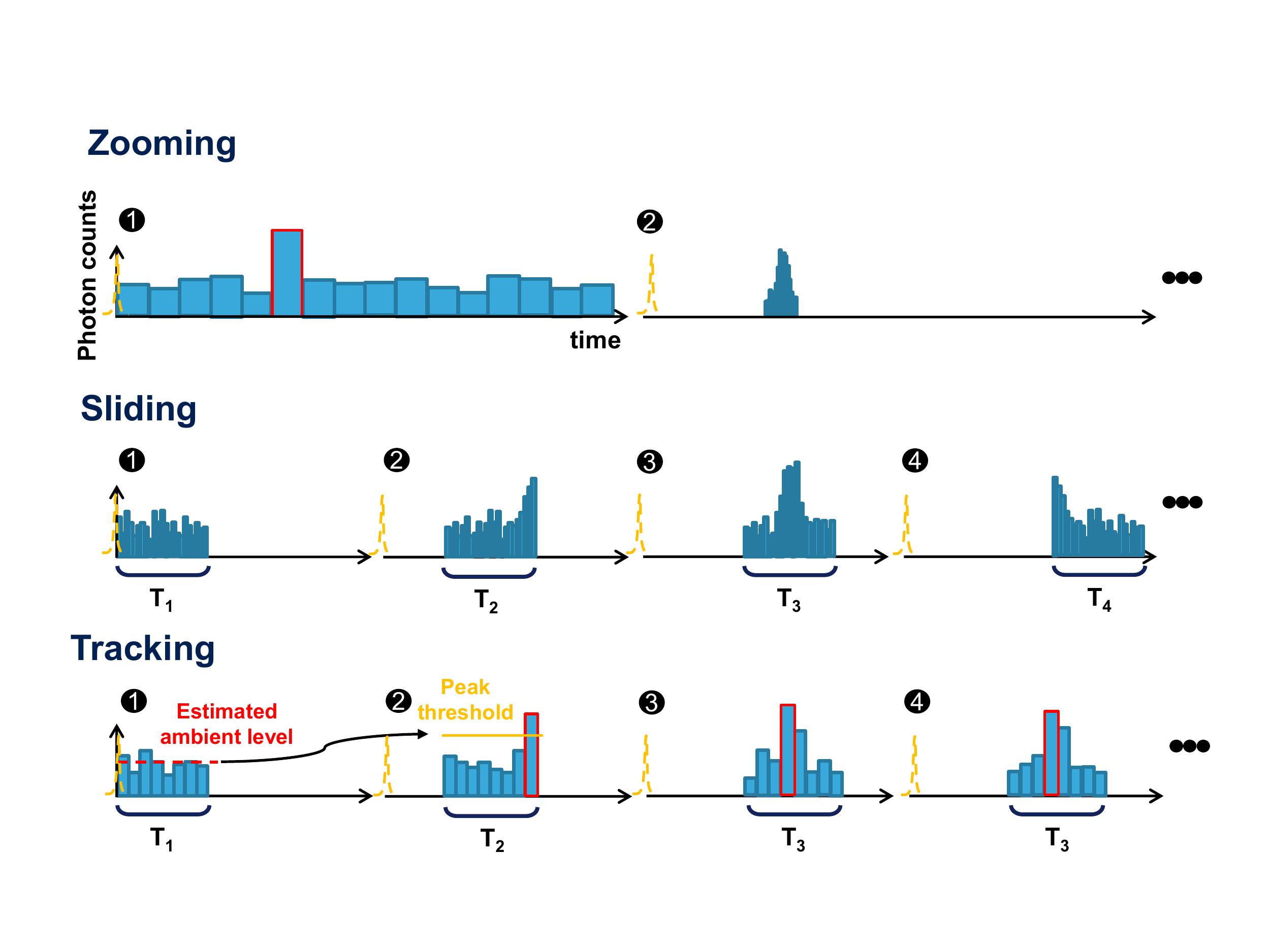}
\caption{Partial histogramming techniques in single-photon direct time-of-flight imaging. In each step of the zooming approach (top plot), the peak bin is identified, and the histogramming logic zooms into the corresponding time range. The sliding approach (middle plot) sweeps through the entire time range in multiple steps. Unlike the zooming approach, the histogram bin width remains fixed. The peak tracking scheme (bottom plot) further develops the sliding concept by estimating the background (baseline photon) level, and detecting signal peaks (marked in red), which are subsequently tracked. The dashed yellow curves correspond to the firing of the laser pulse. Adapted from \cite{Gyongy2022}.
}
\label{fig:fig_partial}
\end{figure}

It has therefore become common to apply multi-step, partial histogramming approaches instead, which typically use just 16 or 32 bins. The approaches, which can be categorised into ``zooming'' and ``sliding'' (Fig.~\ref{fig:fig_partial}), are contrasted in detail in \cite{taneski2022}. The zooming method starts by histogramming the whole time range (using a limited number of bins), and then in each step the peak bin is identified, and the histogramming logic zooms in on the corresponding time range, homing in on the signal peak. In the sliding approach, the partial histogram is swept through the full time range, covering the entire time range over multiple steps. 

As demonstrated in \cite{taneski2022}, the sliding scheme is more effective than zooming over longer ranges and under high ambient levels, due to a large build up of background counts in the initial steps of zooming (resulting from the wide bin width), making it difficult to detect the signal peak reliably, unless very high laser powers are used (in combination with a suitably high histogram bit depth).

The present chip uses the sliding approach, but rather than continually scanning the whole time range, we designed pixels that lock onto peaks and track them. As a result, there is an increase in the effective temporal aperture for signal photons, and once a pixel has homed in a peak, the remaining time range is ignored, precluding the possibility of false peak detection in the excluded range. Furthermore, the approach also provides a mechanism for detecting motion in the observed scene. Read out options include a modality where only pixels detecting a change in depth are reported, demonstrating a pathway towards 3D imaging sensors that not only produce point clouds, but identify spatio-temporal features in the data to reduce data rates, saving power and facilitating low latency vision. 

\begin{figure}[htpb]
\centering
\includegraphics[trim=0.5cm 3cm 0.5cm 3cm, clip=true, width=\linewidth]{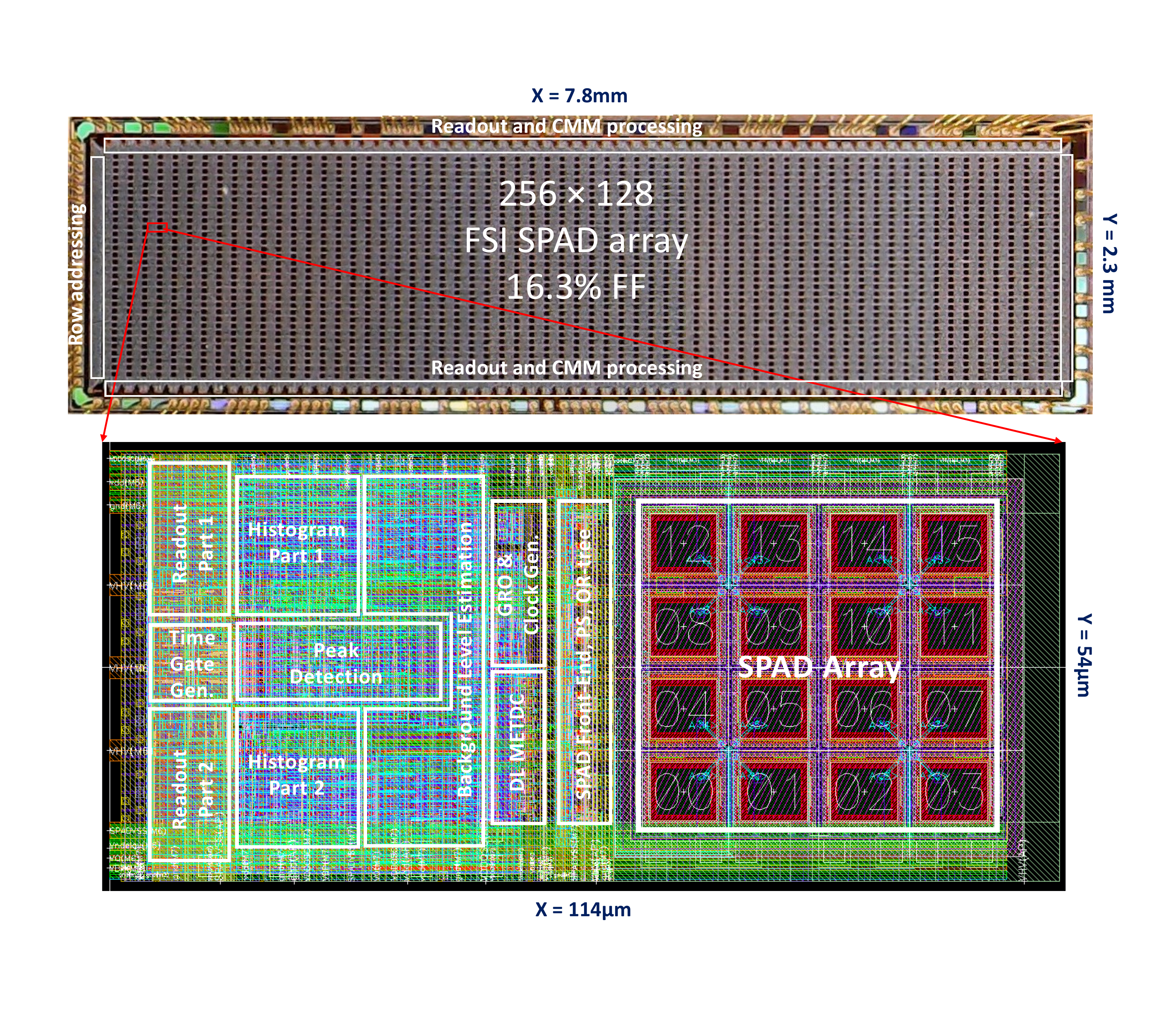}
\caption{Micrograph of the sensor and a close-up of the layout of an individual macropixel, showing the $4\times4$ array of SPADs, and the processing unit next to it. The main constituent processing blocks are framed in white. The chip was implemented in a front side illuminated (FSI) architecture and has an overall fill factor (FF) of 16.3\%.
}
\label{fig:fig_micro}
\end{figure}

\section{Chip Architecture}

The imager is implemented in a standard 40nm process and features a $64\times32$ array of macropixels (Fig.~\ref{fig:fig_micro}). Each macropixel is composed of $4\times4$ SPADs plus a processing unit, which occupy matching areas (of approximately 57$\mu$m $\times$ 54$\mu$m), to enable a potential future 3D stacked implementation \cite{abbas2016}, where the processing units are placed underneath the detector array for increased pixel density and fill factor. 

When depth imaging, each macropixel generates an 8-bin photon timing histogram that is automatically shifted in time to locate and track peaks, via an electronic time gate that is controlled by the macropixel itself. There is additional, and optional, processing outside the array for computing depth values from the histograms with sub-bin resolution using a centre-of-mass (CMM) technique. 

The sensor can also be operated in an intensity imaging modality, where SPADs are grouped in pairs, with each pair reporting a 12-bit photon count for an overall image resolution of $128\times128$.

Data is read out from the sensor over 64, 100~MHz output lines, resulting in maximal frame rates in the 10~kFPS or 100~kFPS range, depending on the mode of operation.

\subsection*{Macropixel architecture}

\begin{figure*}[htpb]
\centering
\includegraphics[trim=1.5cm 5cm 0.5cm 4cm, clip=true, width=0.85\linewidth]{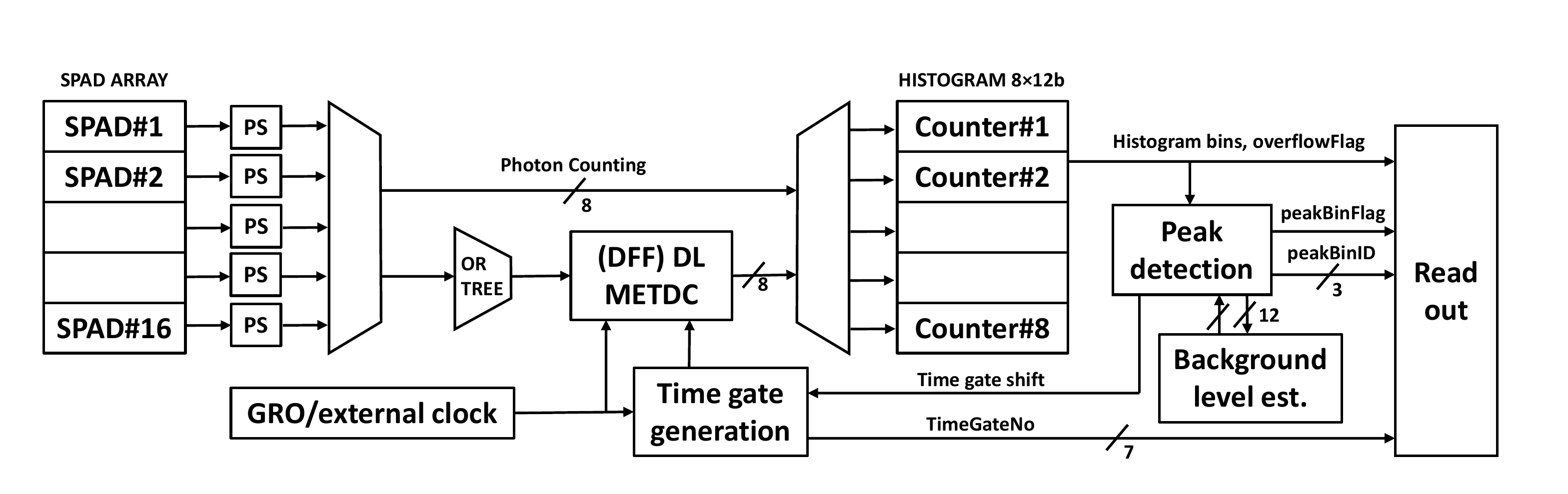}
\caption{Block diagram of the macropixel. When photon timing, the multi-event time-to-digital converter (METDC) uses either a delay line (DL) or a series of delay flip flops (DFF), clocked by an internal ring oscillator (GRO) or external reference clock. For time gate generation, either the GRO or the external clock is used. The ``background estimation'' module also identifies the histogram bin with the maximum count (``peakbinID''), and the ``peak detection'' module calculates the peak threshold and applies it to the histogram. If a peak is detected then the ``peakBinFlag'' and the time gate adjusted according to the location of the peak (``peakbinID''). In case one of the histogram bin counters reaches full capacity (4095), all the bins are frozen, and the ``overFlag'' is raised.
}
\label{fig:fig_macro}
\end{figure*}

A block diagram of the macropixel is shown in Fig.~\ref{fig:fig_macro}. The outputs of the 16 SPADs are fed to pulse shorteners, which produce digital pulses of width $\approx300$~ps for each photon event. In the photon timing modes, the SPAD outputs are then combined using an OR tree and sent to a multi-event time-to-digital converter (TDC) \cite{dutton2015}. The TDC is able to register multiple photons per laser cycle, even if they fall within the same timing bin (provided that the events are sufficient far apart in time to form distinct digital pulses). The output of the TDC goes to 8, 12-bit counters representing the histogram bins. At the end of every exposure, the contents of these counters are processed to estimate the background photon level, to detect statistically significant peaks and to move the time gate position accordingly. If a peak is detected in bin 1 or 2, the time gate position is decremented, if the peak is in bin 7 or 8, or no peak is detected, the time gate is incremented. Otherwise (if the peak is in the middle, so in bins 3 to 6), the time gate position is maintained.

Fig.~\ref{fig:fig_partial} illustrates the peak tracking functionality using histograms captured over four consecutive exposures. The pixel is initially at time gate position 1, and the ambient level is estimated at this position. Based on this ambient level, a peak detection threshold is computed according to Poisson statistics. Next, the time gate is shifted to position 2. In position 2, the last bin is found to exceed the peak threshold. As this is an outlying bin, the time gate is moved again to bring the peak into the middle. The peak is now contained in a middle bin, the time gate position is unchanged for exposure 4. Should the peak disappear in a subsequent exposure, the time gate position is first decremented, before scanning in the normal direction is resumed, if a peak is still not detected. In addition to the ambient level being estimated in exposure 1, the estimate is updated every time there is a shift in the time gate position.

When peak tracking, sensor is typically configured so that there is a 50\% overlap between consecutive time gate positions, which with the signal peak occupying 2-3 histogram bins (identified as optimal for peak extraction assuming a Gaussian peak profile \cite{hagen2007gaussian}) guarantees that there is time gate position containing the peak in its entirety. Assuming a sufficiently large return signal, and neglecting false peak detections, the maximum time to converge to a peak can be expressed as $N\times t_{exp}$ where $N$ is the number of time gate positions and $t_{exp}$ is the exposure time.

Instead of peak tracking, it is also possible to scan through time gate positions continually, as well as to assign specific time gates to different macropixels. Up to 128 different time gate positions are available, leading to an effective overall histogram size of 1024 bins per macropixel.

The coarse timing or time gate position can be generated either using an internal ring oscillator (GRO), or there is the option of using an external clock. For the fine timing or histogram bin width, either a ring oscillator, a delay line (DL), or an external clock is used. In the case of GRO and DL timing the resolution (bin width) is adjustable via external voltages, and can be made as small as 250~ps for DL. In practice, higher photon levels can lead to distortion in the GRO timing, making the delay line and external clock the preferred timing options.

We note that thanks to the adjustable bin size, and the ability to prescribe time gate positions, the sensor could also be conceivably operated in a zooming mode (Fig.~\ref{fig:fig_partial}) with suitable external control.

In addition to the histogram modality, there is also a photon counting mode (as indicated in Fig.~\ref{fig:fig_macro}), where SPADs are combined in pairs, and each histogram bin is repurposed to serve as a photon counter for a pair of SPADs. It is possible to switch between the histogram and photon counting modes on the fly (with time gate positions being preserved), for instance for the purpose of off-chip intensity-guided depth upscaling \cite{Gyongy2020}.

\subsection*{In-pixel peak detection}

The existence of a significant peak within the 8-bin histogram captured by a macropixel is determined by comparing the bin with the maximum count $h_{max}$ with the threshold given by the equation
\cite{gnecchi20171}:
\begin{equation}
h_{thresh}=B+1.75\alpha\sqrt{B},
\label{eq:eq_thresh}
\end{equation}
where $B$ is the estimated background level and $\alpha$ can be set to 1 or 2. Thus, under the assumption of Poisson noise on the bin counts, the peak has to be more than 1.75 or 3.5 standard deviations away from the baseline ambient level for it to be considered a ``true'' peak, with $\alpha$ setting the desired level of detection sensitivity.

\begin{figure}[htpb]
\centering
\includegraphics[trim=3.5cm 2cm 5cm 4cm, clip=true, width=\linewidth]{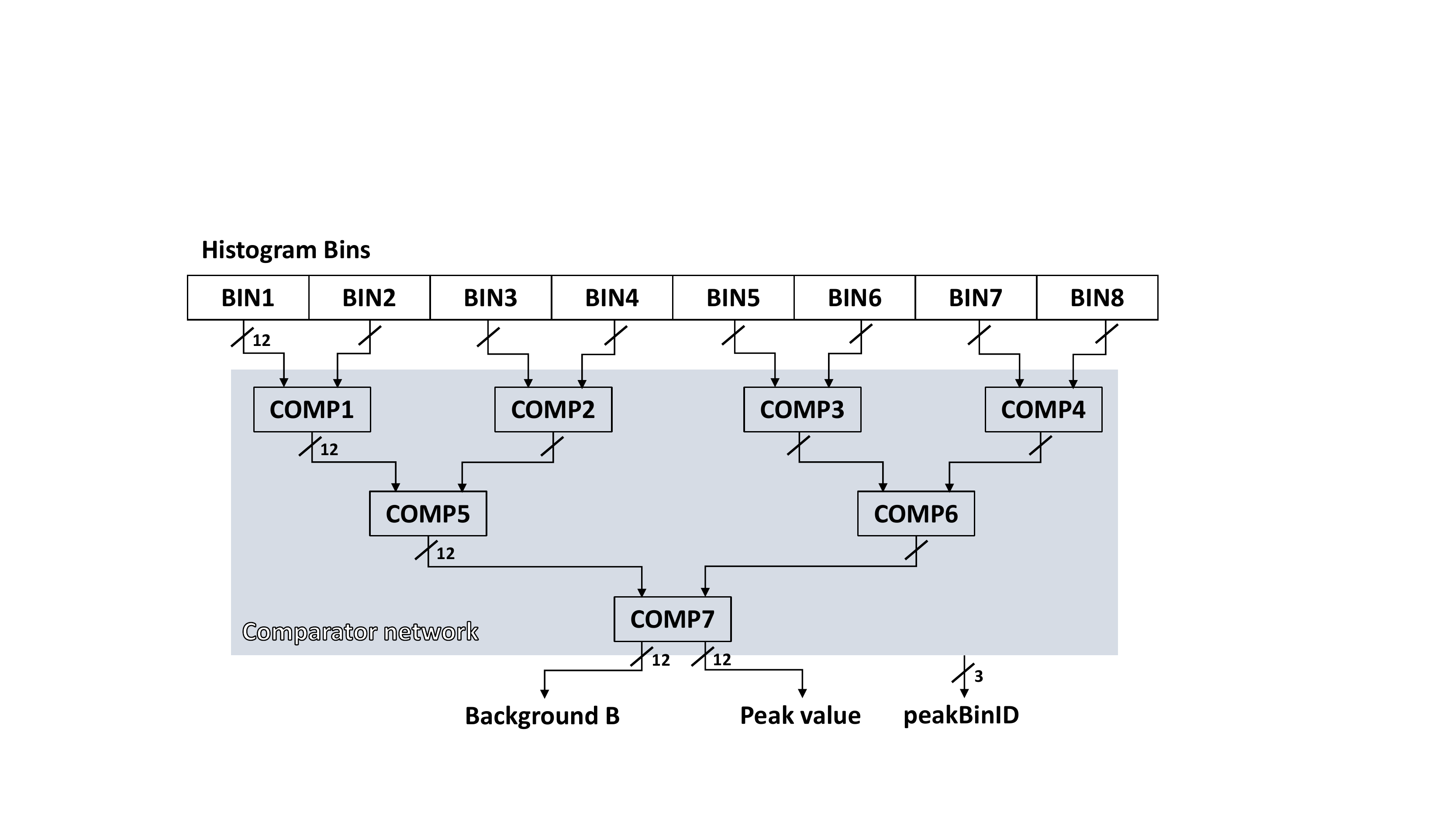}
\caption{Comparator tree used within the macropixel to identify the histogram bin with the maximum count $h_{max}$, and to estimate the background level $B$.
}
\label{fig:fig_peak_bin}
\end{figure}

The bin with the highest count (or ``peak bin'') is identified using the network of comparators shown in Fig.~\ref{fig:fig_peak_bin}, with each comparator returning the higher of its two inputs. To save circuit space, the network is reused in the estimation of the ambient level $B$, which is taken as the lower of the two inputs of the final comparator COMP7. 

\begin{figure*}[htpb]
\centering
\includegraphics[trim=0cm 11cm 0cm 10cm, clip=true, width=\linewidth]{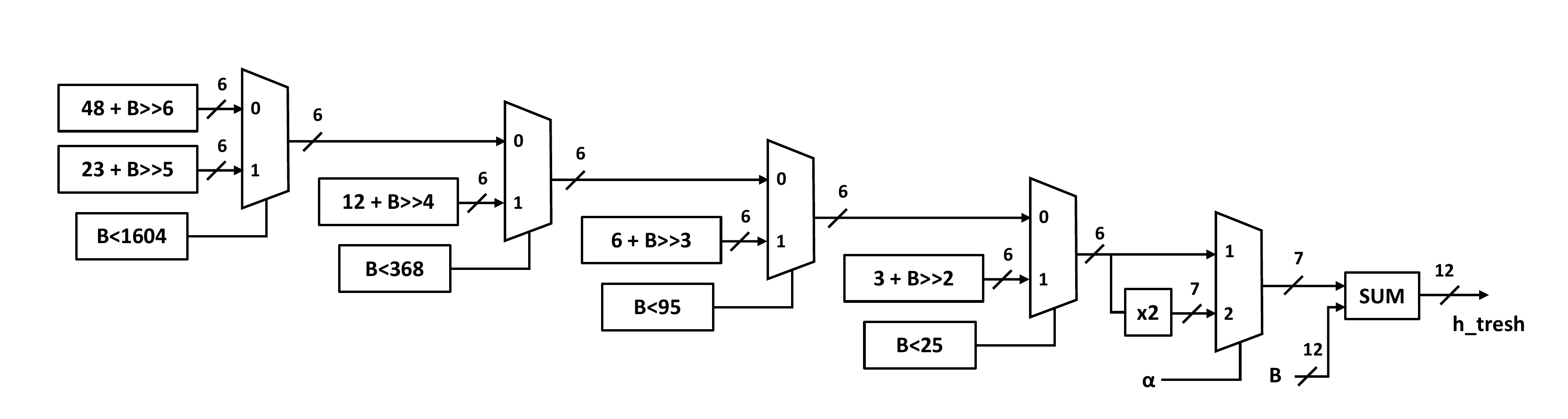}
\caption{Logic used for peak threshold ($h_{thresh}$ ) calculation, based on the estimated background level $B$. The logic includes a piece-wise linear approximation of the quadratic function (see also Fig.~\ref{fig:fig_thresh_calc}). $B$ is compared to a set of values, defining the end points of the linear segments. The linear functions within the segments are implemented via bit shifting (``>>'') and the addition of a constant term (specific to the given segment). The detection sensitivity can be adjusted via the $\alpha$ input.
}
\label{fig:fig_thresh}
\end{figure*}

To calculate the peak detection threshold in eqn.~\ref{eq:eq_thresh}, a piece-wise linear approximation is used for the factor $1.75\sqrt{B}$, as indicated in Fig.~\ref{fig:fig_thresh}, which is implemented using comparator, bit shifting, and summation logic. Each linear segment in the approximation has a gradient in the form of $1/2^n, n=1,2...$ to allow computation using simple bit shift logic. The start and end points of the segments (which are hardwired into the logic) were determined using least-squares optimisation.  

\begin{figure}[htpb]
\centering
\includegraphics[trim=0cm 0cm 0cm 0cm, clip=true, width=0.8\linewidth]{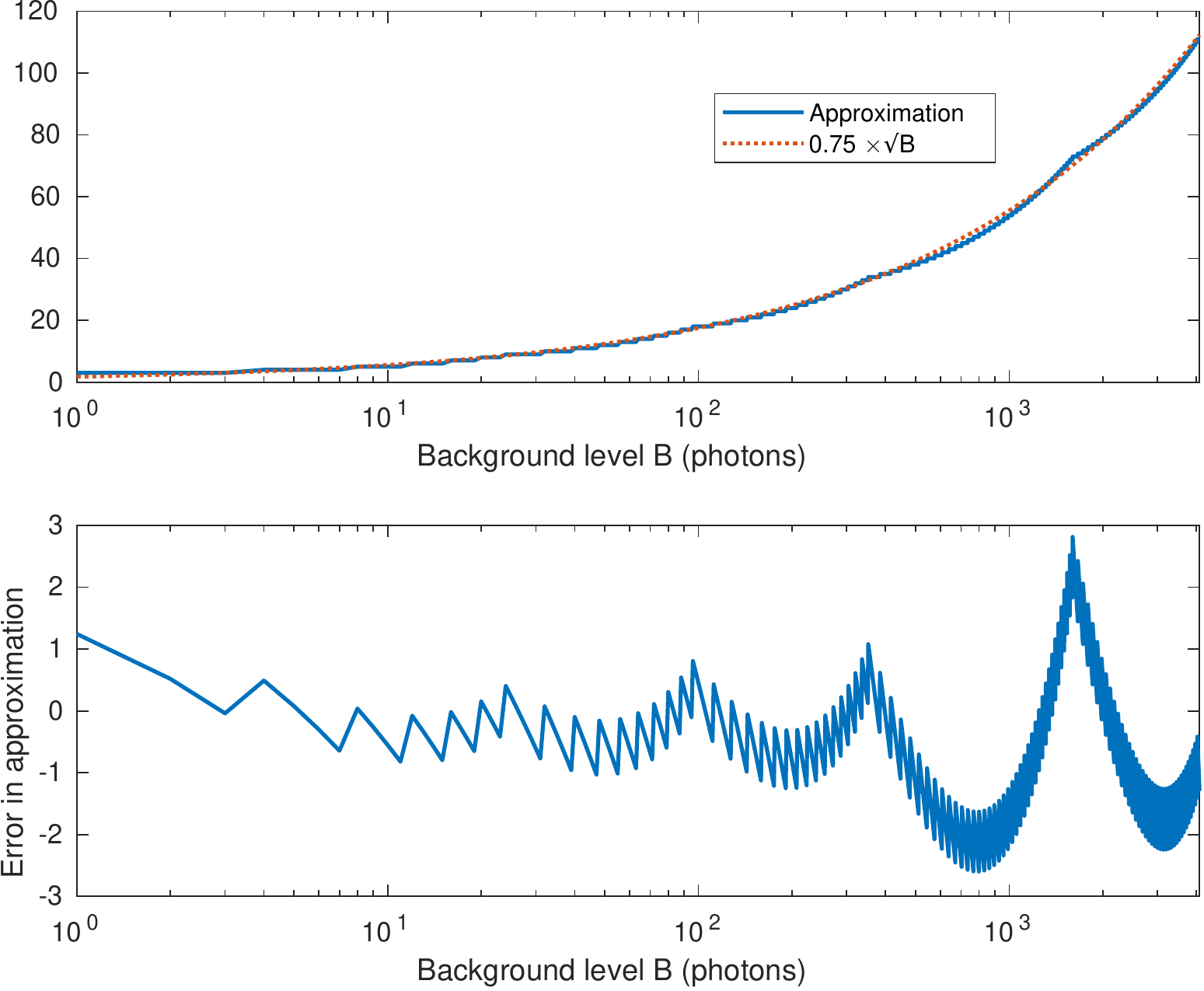}
\caption{Accuracy of the approximation for the quadratic function built into the macropixels for peak threshold calculation. The top graph plots the quadratic function and the approximation for background levels $B$ ranging from 0 to 4095 photons (the maximum possible level that can be recorded). The bottom graph shows the error in the approximation over the same range of $B$.
}
\label{fig:fig_thresh_calc}
\end{figure}

Fig.~\ref{fig:fig_thresh_calc} shows the accuracy of the approximation of $1.75\sqrt{B}$ over the full range of possible background levels $B$. The maximum error is seen to be $<3$ photons. 

\subsection*{Output modes for depth imaging}

In addition to the default histogram output, where all the histogram bins are read out, there are two modes providing further data compression in the form of direct depth readings (Table~\ref{tab:modes}) .

In the bin-resolution depth mode, each macropixel reports the position of the bin with the maximum count (and whether this is a statistically significant peak). An almost tenfold reduction is thereby achieved in the amount of data read out per frame.

In the sub-bin resolution depth mode, peak extraction is performed outside the array (in column parallel logic) using a centre-of-mass (CMM) approach, which again results is a significant ($\times7$) data compression compared with reading out histograms.

In addition to the modes in the table, there are two smart readout options, where only macropixels with a peak or those with a peak that is moving are read out, and the rest are replaced by zeros. A moving peak is defined here as one requiring a shift in time gate position. These selective readout options aim to reduce toggling on the output lines as well as downstream processing requirements. The latter option where only pixels with changing depth are read out has parallels with dynamic vision sensors \cite{delbruck2010}, although the readout is frame-based here rather than asynchronous, event-based. 

Primarily intended for low-ambient environments, there is the further option of applying multiple subexposures between frame readouts to macropixels that are peak-searching, to allow for faster convergence to peaks. Furthermore, during readout, rows can be selectively skipped, via external control, for example if none of the macropixels in a row have detected a peak.

\begin{table}[htbp]
\centering
\caption{\bf Comparison of the different output formats offered by the sensor for depth imaging, in terms of the type of output data, the number of bits reported per macropixel, and the maximum achievable frame rate.}
\begin{tabular}{cccc}
\\[-0.5em]
\hline
Mode & \begin{tabular}{@{}c@{}}Output \\[-0.25em]format\textsuperscript{*}\end{tabular}  & \begin{tabular}{@{}c@{}}\# bits per \\[-0.25em] macropixel\end{tabular}
  & \begin{tabular}{@{}c@{}}Max. frame \\[-0.25em]rate (FPS)\end{tabular} \\
\hline
Histogram & \begin{tabular}{@{}c@{}}hist bins and\\[-0.25em]hist peak data\end{tabular} & 108 & 29k \\
Bin res. depth & hist peak data & 12 & 260k \\
\begin{tabular}{@{}c@{}}Sub-bin\\[-0.25em]res. depth\end{tabular} & \begin{tabular}{@{}c@{}}centre of mass\\[-0.25em]of hist bins\textsuperscript{**}\end{tabular}  & 15 & 208k \\

\hline
\\[-0.75em]
\multicolumn{4}{l}{{\small \begin{tabular}{@{}l@{}}\textsuperscript{*}in addition to 7-bit time gate position; the histogram peak data\\[-0.25em]consists of peak bin flag, peak bin ID and overflow flag\\\textsuperscript{**}after compensation for background counts\end{tabular}}}\\
\end{tabular}
  \label{tab:modes}
\end{table}

\subsection*{On-chip depth computation using CMM}

For computing a sub-bin resolution estimate for the depth $d$, we use a variant of the centre-of-mass scheme \cite{Gyongy2020}, described by the following equation:
\begin{equation}
\hat{d}=\frac{\sum_{t=1}^{8}t\left(h_t -B\right)}{\sum_{t=1}^{8}\left(h_t -B\right)},
\label{eq:cmm}
\end{equation}
where $h_t$ $(t=1...8)$ are the histogram bins at a given macropixel, and $B$ is the estimated background level. To simplify the hardware implementation, $B$ is taken here as the minimum bin count amongst the 8 bins.

\begin{figure}[htpb]
\centering
\includegraphics[trim=2cm 8cm 2cm 7cm, clip=true, width=\linewidth]{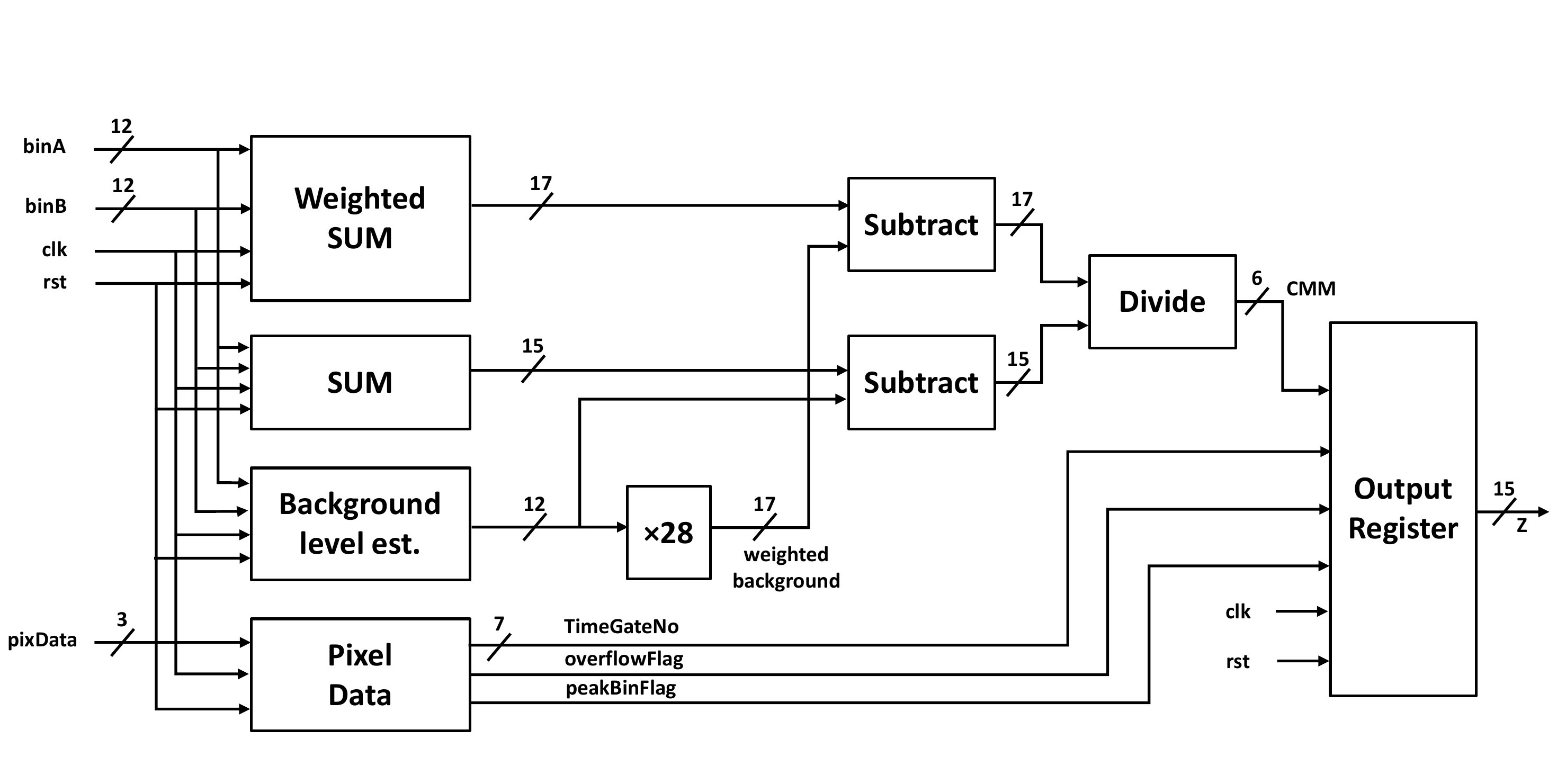}
\caption{Block diagram of on-chip centre-of-mass (CMM) processing module providing sub-bin resolution depth estimates.
}
\label{fig:fig_cmm}
\end{figure}

The implementation of the CMM module is shown in Fig.~\ref{fig:fig_cmm}, with the division being carried out by a parallel divider to a three decimal point accuracy, allowing the position of the peak to be determined with a resolution of $1/8$ of the bin size. The bin counts are loaded into the module two at a time. Due to a design bug affecting the current version of the chip, there is an error in the order in which the bins are presented to the module. As a consequence, the histogram peak has to be contained in the middle two bins for $\hat{d}$ to be correctly calculated.

\section{Results}

To create a camera system, the sensor is attached to a custom PCB board alongside an Opal Kelly XEM7310 FPGA integration module, which generates the required control signals and streams the SPAD data over a USB~3.0 link to a PC running a Matlab software interface. A 25~mm/f1.4 objective (Thorlabs MVL25M23) is used in front of the sensor, giving an approximately 20$\times$5~degree field-of-view (FOV). For depth imaging, the sensor is paired with a compact 850~nm VCSEL source with 10~ns pulse width and 60~W peak optical power triggered at 1.2~MHz. Lenses are used to shape the laser beam according to the sensor FOV, and a 10~nm ambient filter (Thorlabs FL850-10) is added to the receiver optics. The whole system is powered using standard USB ports: a USB~3.0 port supplies the camera board, and a USB-C port powers the laser.

\subsection*{Intensity imaging}
\begin{figure}[htpb]
\centering
\includegraphics[trim=3cm 11cm 3cm 11cm, clip=true, width=0.9\linewidth]{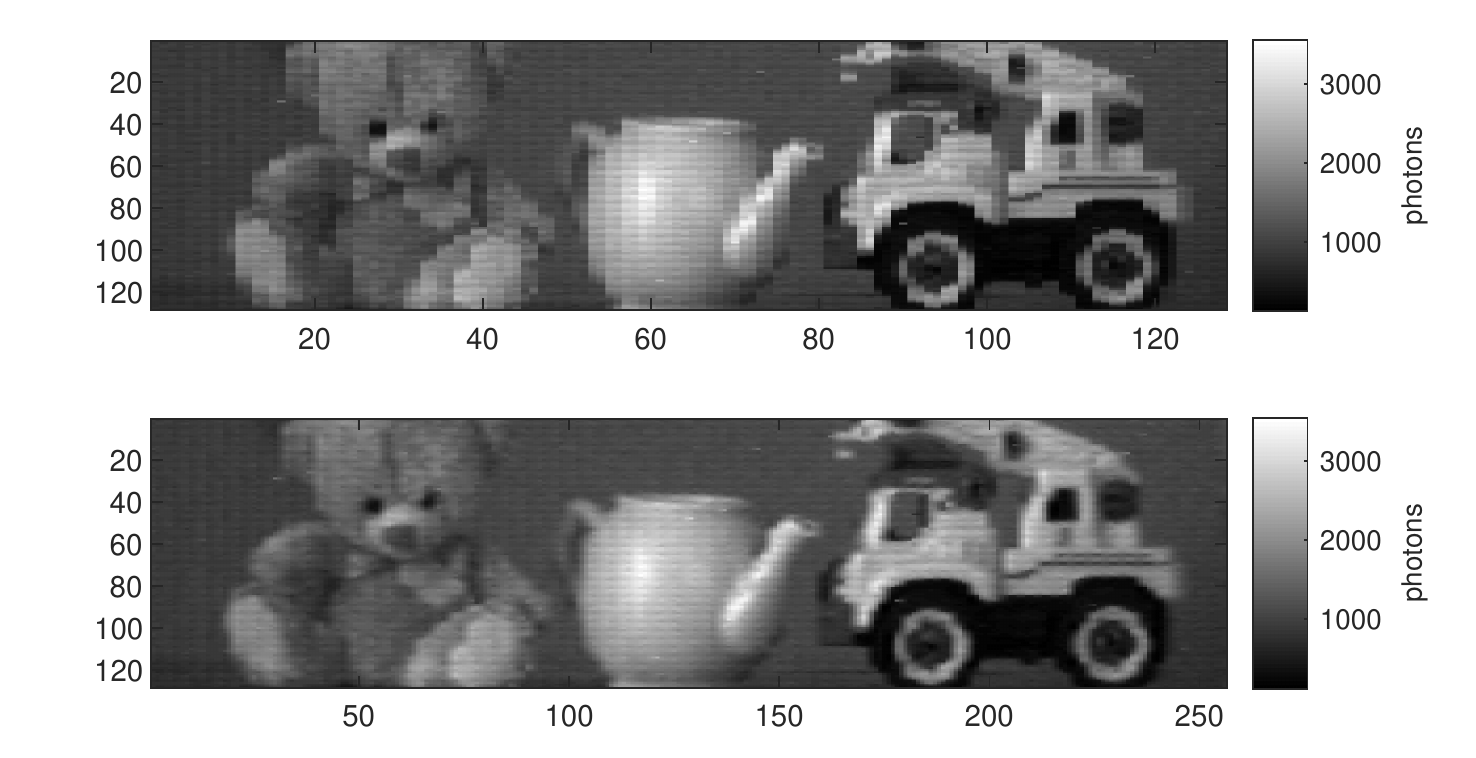}
\caption{Example photon counting (intensity) image. The top image depicts the raw data from the sensor, the bottom image is the result of interpolation to account for the uneven spacing of pixels horizontally. The data was captured indoors with a 200~$\mu$s exposure time, equating to 5~kFPS.
}
\label{fig:intensity}
\end{figure}

Fig.~\ref{fig:intensity} shows an example photon counting image captured by the sensor. The raw image (top image) is seen to be pixelated horizontally, due to the uneven spacing of pixels (as a result of the gaps created by the processing unit in each macropixel). By accounting for this uneven spacing, and applying interpolation, the level of pixelation is reduced (bottom image).

\subsection*{Histogram non-linearity}
\begin{figure*}[htpb]
\centering
\includegraphics[trim=3.5cm 6.5cm 4cm 5cm, clip=true, width=0.9\linewidth]{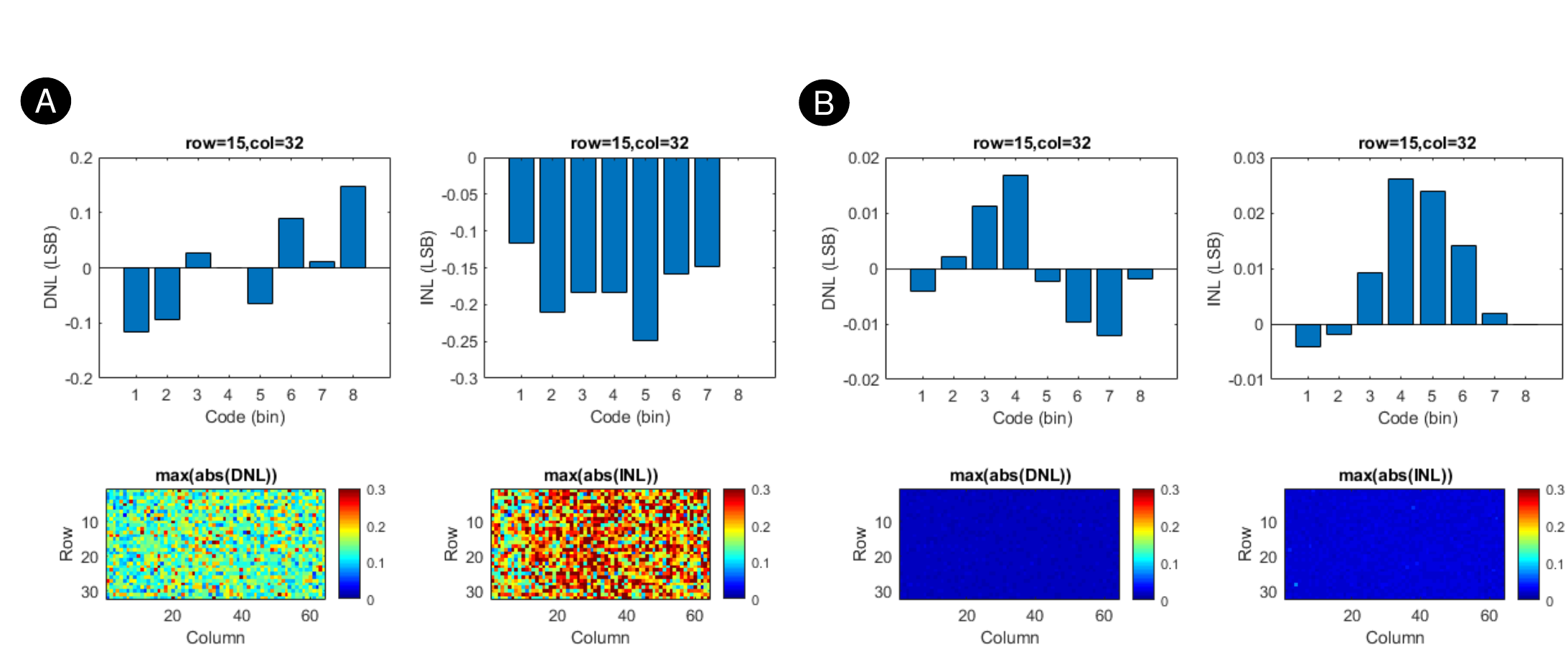}
\caption{Non-linearity in the photon timing histogram: a) for a nominal bin width of 1~ns, when the delay line is used for timing, b) for a nominal bin width of 8~ns,  when an external clock (of frequency 125~MHz) is used for timing. The top figures plot the differential and integral nonlinearity (DNL/INL) for a selected pixel; the bottom figures show the maximum DNL and INL across the pixel array.
}
\label{fig:nonlin}
\end{figure*}

The non-linearity of the sensor when photon timing was measured by capturing a large number of exposures under ambient conditions, and assessing the bin-to-bin differences in the mean photon count per bin within each macro pixel. Figs.~\ref{fig:nonlin}a and b show example results for the delay line and external clock timing options, respectively, each based on 4000 exposures. For delay line fine timing with a 1~ns bin size the maximum DNL for a typical (median) pixel is found to be $\approx$14\%. In the case an external clock is used for timing, a nonlinearity of $\approx$1.6\% was measured for a 8~ns bin size (the minimum bin size obtainable in this configuration being subject to bandwidth limitations in feeding a clock into the chip and distributing it across the array).

Due to the superior linearity of external clocking timing, and the pulse width of the compact laser sources used here being better matched to wider bin sizes, the external clock option was adopted as a default configuration, and depth imaging results presented here were obtained with an external clock.

\subsection*{Specificity of peak detection}

\begin{figure}[htpb]
\centering
\includegraphics[trim=0cm 0cm 0cm 0cm, clip=true, width=0.6\linewidth]{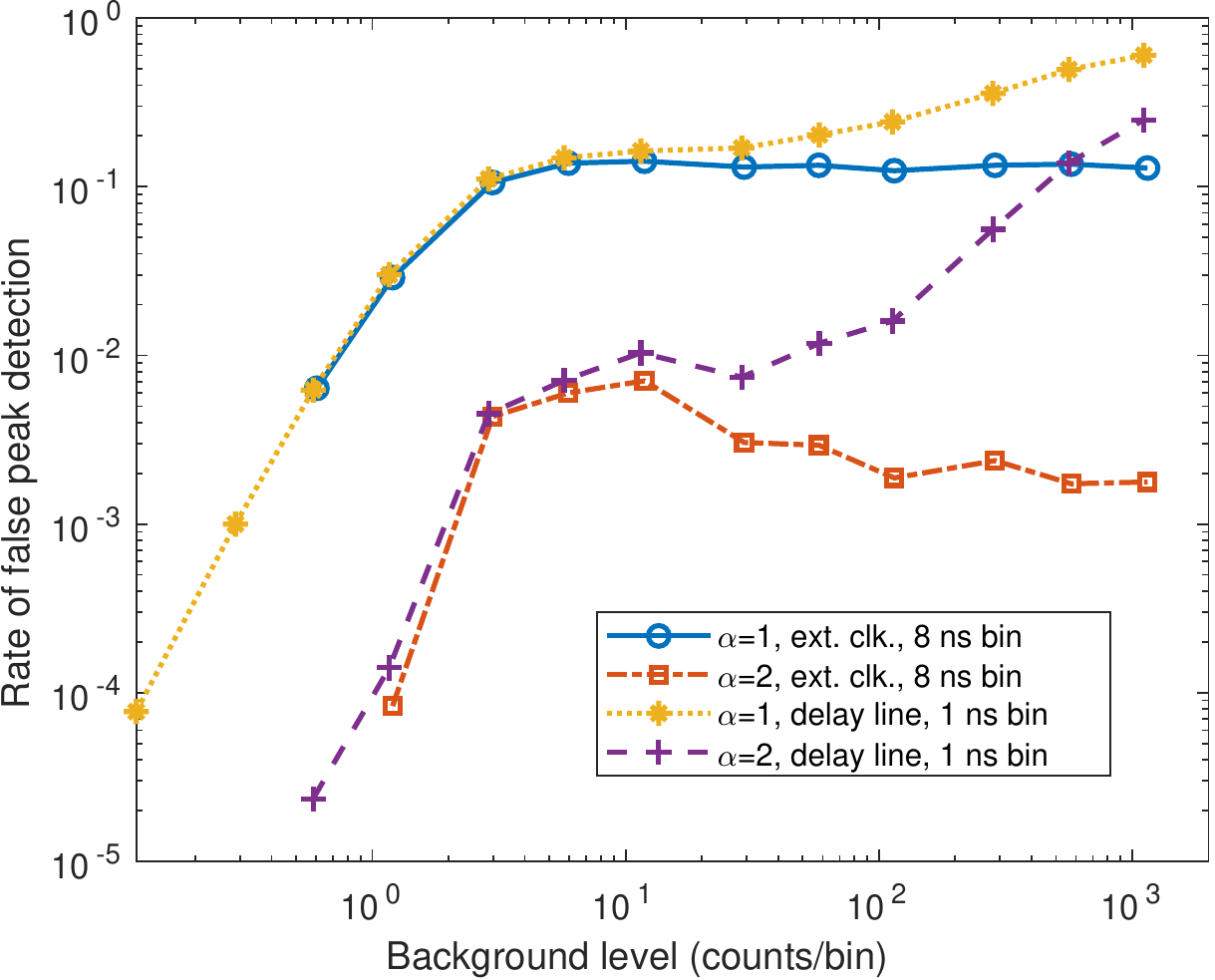}
\caption{False positive rate in the peak detection versus the background level for different $\alpha$ settings and timing modes. Each data point was generated using 500 frames, and corresponds to the mean rate of false positive detections across the array. The background level was measured as the mean bin count and varied by changing the exposure time.
}
\label{fig:fpr}
\end{figure}

The specificity of the in-pixel peak detection logic was tested by recording sequences of frames under ambient light. The measurements were repeated for different background levels by varying the exposure time. Fig.~\ref{fig:fpr} plots the resulting false positive rate for the two different sensitivity settings, in the case of both external clock and delay line timing. For external clock timing with $\alpha=1$, the false positive rate (FPR) is around 10\% at higher background levels; for $\alpha=2$ the FPR approaches 0.2\%. It is noted that these values are higher than what would be expected in theory assuming a known background level $B$ (4\% and 0.023\% for z scores of 1.75 and 3.5 \cite{larsen2005introduction}) due to the approximate nature in which $B$ is estimated in-pixel. For delay line timing, the FPR continues to climb as the background level is raised, due to the higher non-linearity in this mode. To ensure an FPR of, say <2\%, the background level should be around $\approx$100 counts/bin or lower.

\subsection*{Short range, indoor imaging}

\begin{figure}[htpb]
\centering
\includegraphics[trim=5cm 3.5cm 5cm 2.5cm, clip=true, width=\linewidth]{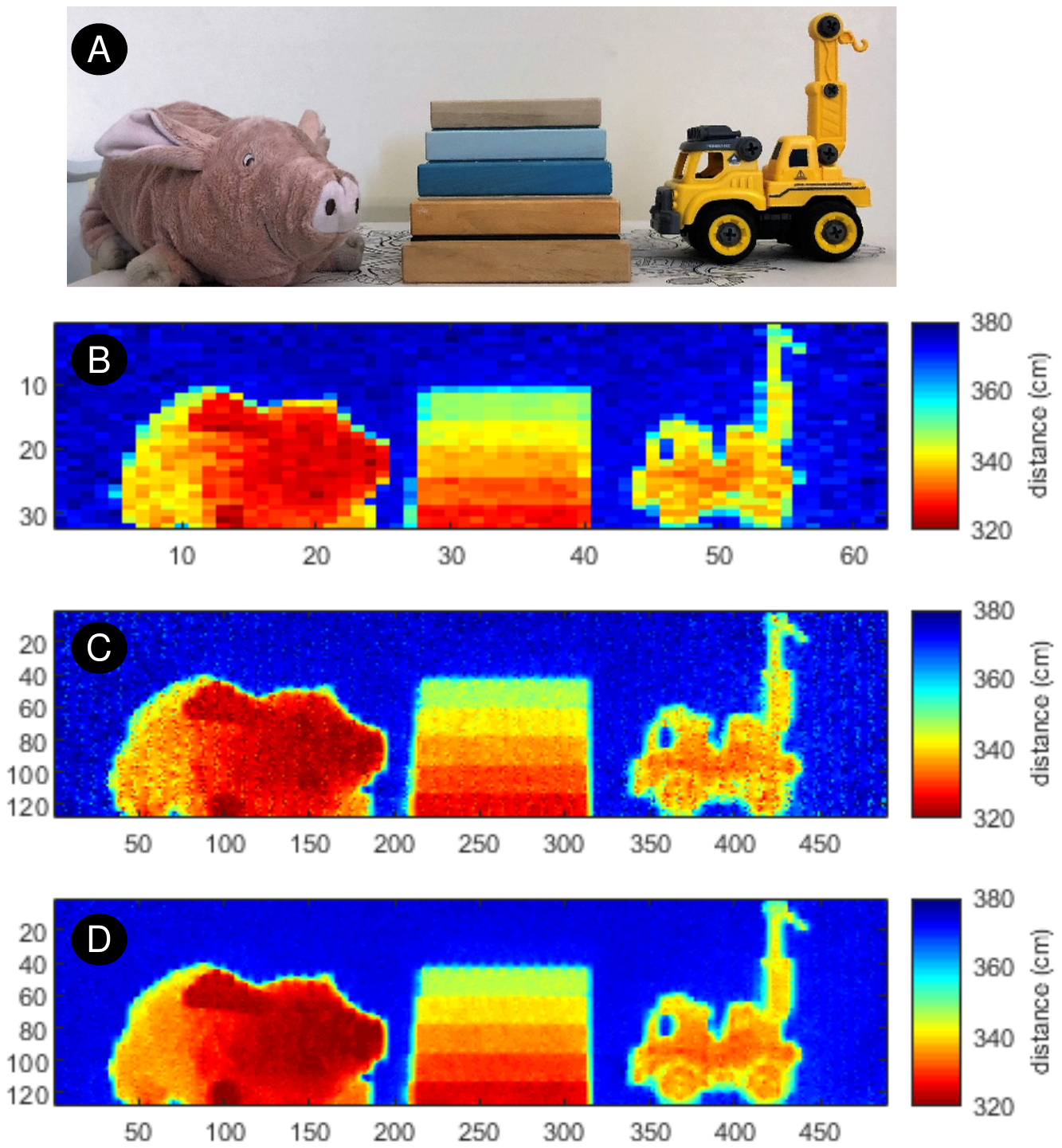}
\caption{3D-imaging results for indoor scene with toys: a) RGB image of the scene; the separation between the wooden blocks is $\delta z\approx5$~cm, b) depth map from a single 1~ms exposure (1~kFPS frame rate), c) the result of $16\times3.2$~ms exposures being combined off-chip for a high resolution depth map (20~FPS overall frame rate), d) the average of 20 high resolution images. To produce the higher resolution depth map, one SPAD at a time was enabled within the 4x4 array in each macropixel, the vertical dead zones between the pixel columns being interpolated over in post-processing.
}
\label{fig:shortrange}
\end{figure}

For short range imaging, there is no need to scan across time gate positions, as a single time gate provides a 9.6~m unambiguous distance range (assuming external clock timing). The time gate is thus fixed at the first position. Fig.~\ref{fig:shortrange} shows example results obtained indoors with a fixed time gate. Fig.~\ref{fig:shortrange}b depicts the depth map from a single exposure in the histogram readout mode, with depth extraction carried out off-chip \cite{Gyongy2020}. As previously demonstrated in \cite{stoppa2021}, it is possible to increase the lateral resolution of a macro-pixel-based imager by successively enabling individual SPADs within the macropixel. A similar time-multiplexing approach can be applied to the present sensor, with multiple exposures being taken and combined off-chip to compose a higher resolution depth map (bottom image in Fig.~\ref{fig:shortrange}c and d).

\begin{figure}[htpb]
\centering
\includegraphics[trim=4cm 6cm 5cm 6cm, clip=true, width=0.9\linewidth]{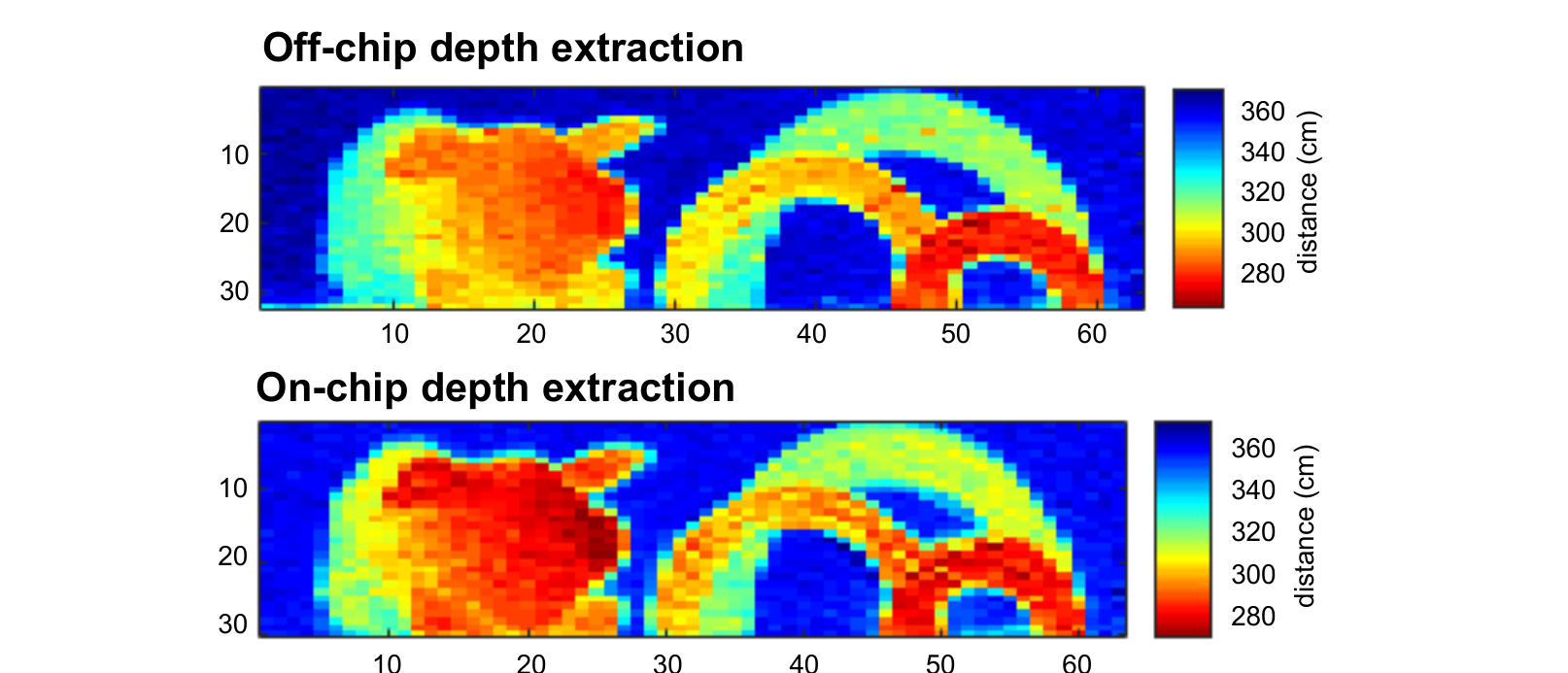}
\caption{
3D-imaging results for indoor scene with plush pig and wooden arcs. The top image corresponds to off-chip depth extraction (histogram readout), the bottom image is for on-chip (CMM) depth computation. Both images were captured with a 1~ms exposure time (1~kFPS).
}
\label{fig:cmmimage}
\end{figure}

Another short range example is given in Fig.~\ref{fig:cmmimage}, which compares two depth maps of the same scene, the top image having been obtained in histogram mode (with off-chip peak extraction) and the bottom image being a result of on-chip (CMM) depth computation. 
Recalling the bug currently affecting CMM, an exact correspondence cannot be expected. Nevertheless, the two depth maps are visually similar. Noting that the depth in the scene varies by a distance equivalent to less than a single histogram bin, the results demonstrate the ability of CMM to output sub-bin resolution readings.

\subsection*{Medium range, outdoor imaging}

\begin{figure}[htb]
\centering
\includegraphics[trim=0cm 0.5cm 0cm 0cm, clip=true, width=0.8\linewidth]{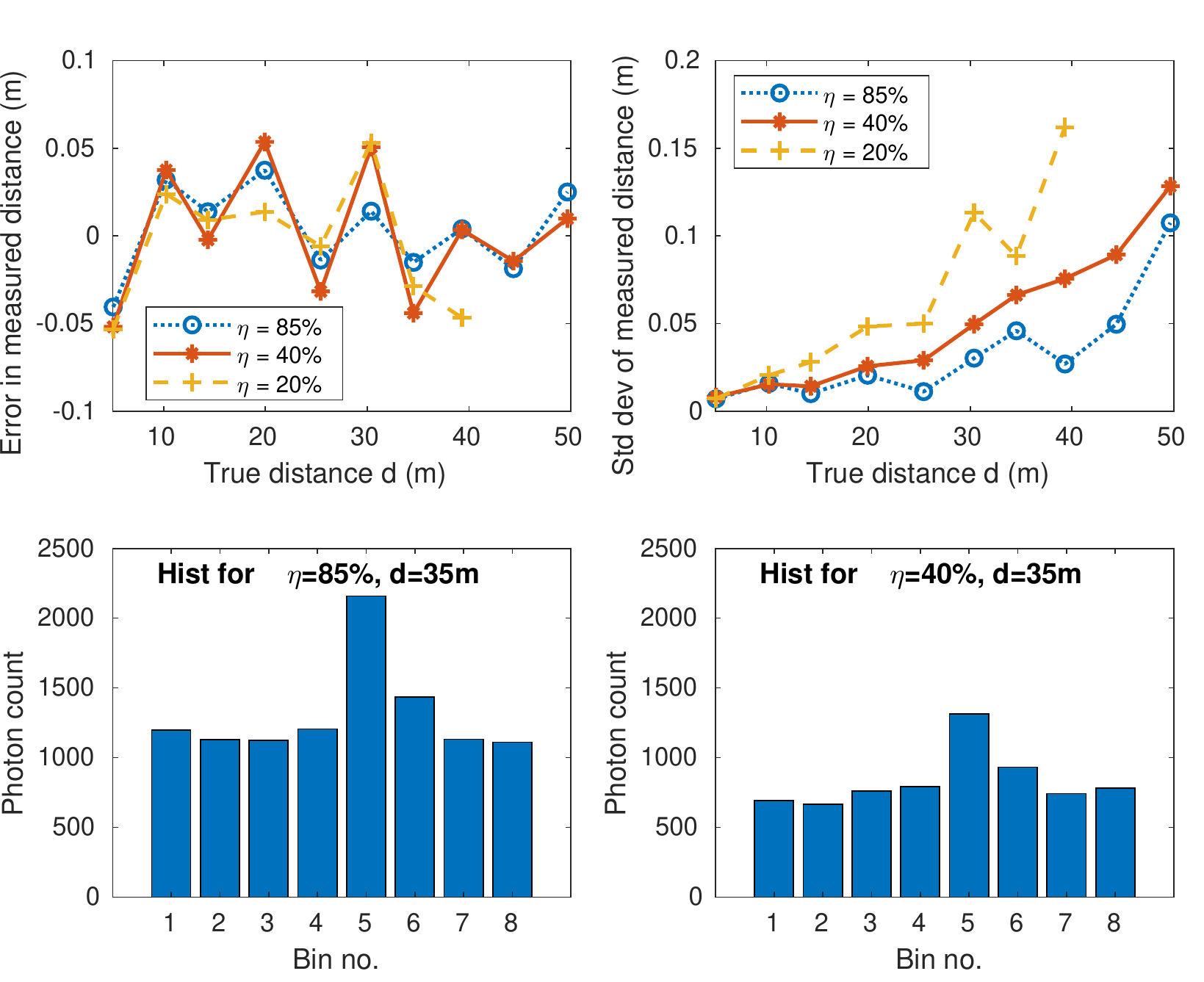}
\caption{
Characterisation of system accuracy and precision for two different target reflectivities over a 50~m distance range. The top graphs plot the error and standard deviation in the measurement. Each data point corresponds to a single macropixel observing the target, and is computed from 168 consecutive exposures. The bottom graphs show sample histograms for the two targets at a distance of $d=35$~m. Note that at $d=45$~m and 50~m the observed target width became less than the size of the macropixel, resulting in an increased level of background photons relative to signal photons. A target of 20\% reflectivity was also used in the test and was reliably detected up to $d=40$~m, where the standard deviation was 16~cm, and the histograms showed a signal-to-background ratio of 0.04.
}
\label{fig:sweep}
\end{figure}

For imaging outdoors the peak tracking functionality was activated with $N=16$ time gate positions (giving 81.6~m of unambiguous range) and $\alpha=2$. The accuracy and precision of the system were evaluated for flat targets of different reflectivity $\eta$ up to range of 50~m. Ground truth readings were taken with a laser range finder (Bosch GLM 250 VF), and the ambient level was measured with a calibrated light meter to be 20~kLux. Fig.~\ref{fig:sweep} summarises the results for $\eta$=0.4,0.85, obtained at a frame rate of 50~FPS. The results indicate a maximum non-linearity of 5~cm and a precision (standard deviation) of <13~cm. A target detection rate of 100\% was measured throughout. In practice, the performance of the system is dependent on the laser source that the sensor is paired with, and given the availability of laser sources with much higher peak power than the 60~W VCSEL source used in this study, we expect longer ranges, under potentially higher ambient levels, to be achievable.

\begin{figure*}[htpb]
\centering
\includegraphics[trim=1.5cm 9cm 2cm 6cm, clip=true, width=\linewidth]{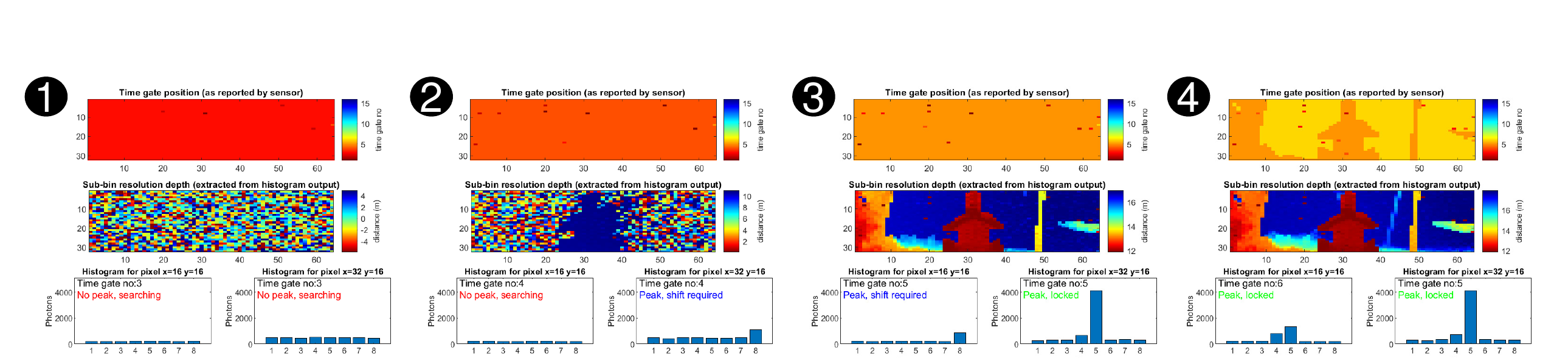}
\caption{Frames from sequence with person holding a ball, captured at 50~FPS. The top graphs show the time gate position, the middle graph is the extracted depth, and the bottom graphs show histograms from two selected macropixels. The panels correspond to the third exposure onwards following switch on.}
\label{fig:sequence}
\end{figure*}

The peak tracking functionality is illustrated in Fig.~\ref{fig:sequence} through a sequence of consecutive exposures after switch on. A panel is shown for each exposure, plotting the time gate position (top graph), the depth extracted from the histograms (middle graph), and the histograms corresponding to two macropixels (bottom), one in the left side of the field-of-view and the other in the centre. In the first exposure (panel 1) no surfaces can be seen, and both of the highlighted macropixels move to the next time gate position. After exposure 2, a surface starts emerging in the captured scene. The second macropixel has detected a peak but it is at an outlying bin, so a further time shift is required. The first macropixel is still peak searching. Exposure 3 reveals most of the scene, which is of a person sitting next to a tree and holding a ball, with a swing on the other side. Macropixel 1, which observes the ball has now locked onto the peak. Macropixel 2, looking at the background has detected a surface but still needs a time shift. After exposure 4, most of the pixels have homed in, including both of the highlighted macropixels. More details have appeared in the scene, including the second leg of a swing. Subsequent exposures (not shown), are mostly unchanged, apart from a few remaining macropixels also converging. 

\begin{figure}[htpb]
\centering
\includegraphics[trim=6cm 4.5cm 6cm 4cm, clip=true, width=\linewidth]{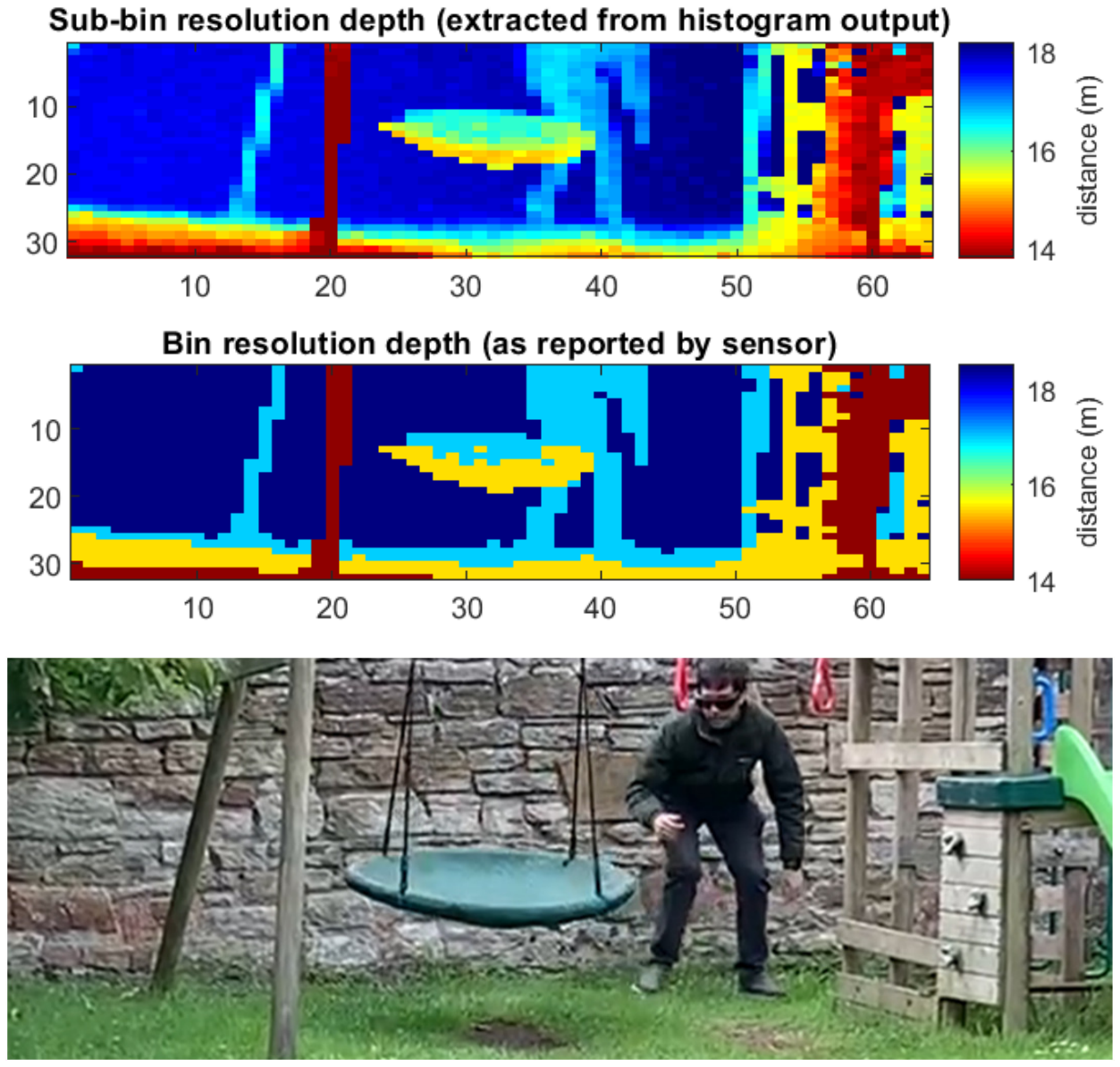}
\caption{
Snapshot of sequence with swing, captured at 50~FPS. The top graph shows the depth extracted from the histograms; the middle graph gives the bin resolution depth (calculated on-chip based on the histogram bins with the highest photon count, with no interpolation applied). In the bottom an RGB image is shown of the scene. See also Visualisations \href{https://youtu.be/qlIr0gX-t2M}{1} and \href{https://youtu.be/6YGAO-eOoWM}{2}.
}
\label{fig:swing}
\end{figure}

\begin{figure}[htpb]
\centering
\includegraphics[trim=5cm 8.1cm 5cm 1cm, clip=true, width=0.9\linewidth]{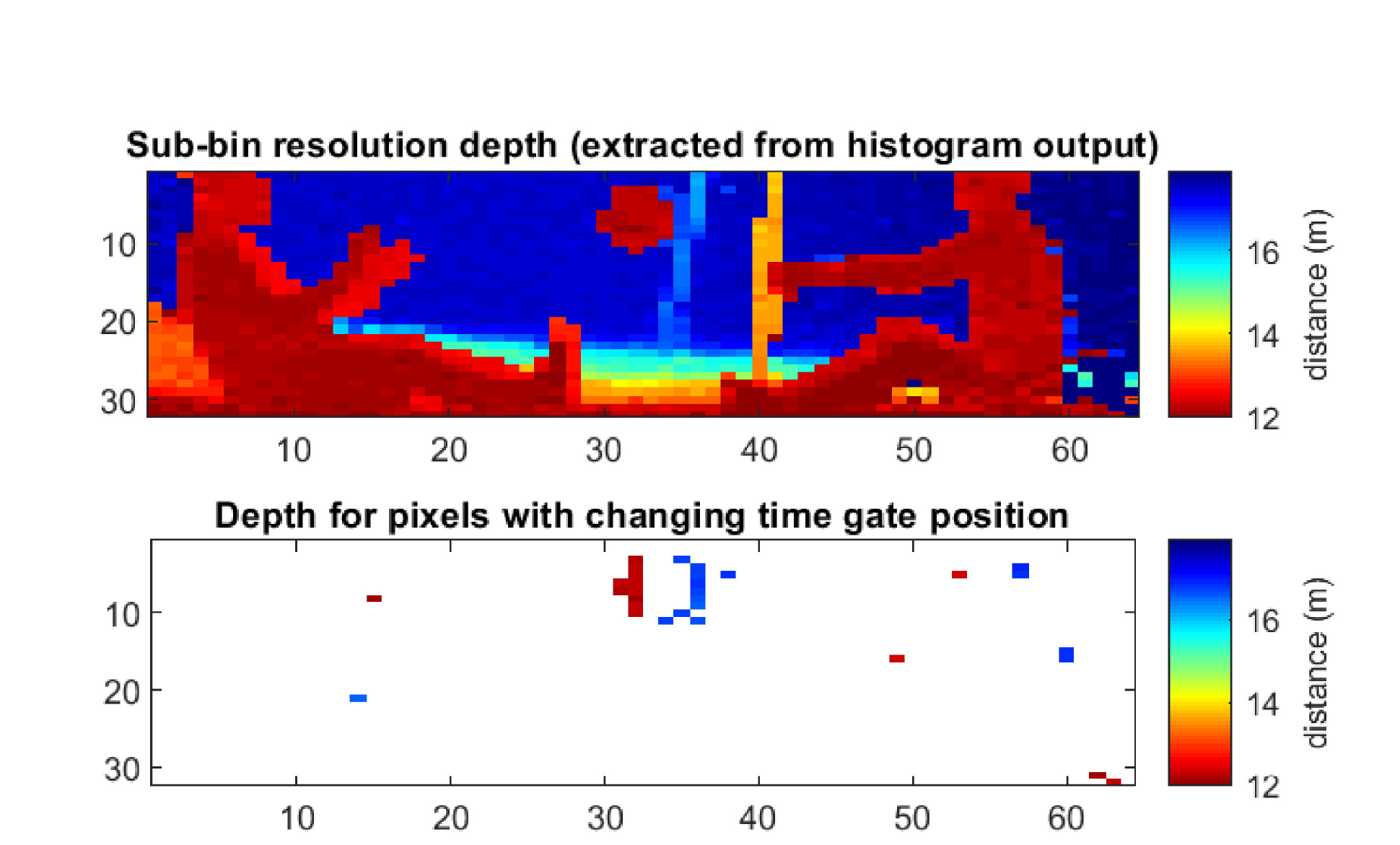}
\caption{
Snapshot of ball passing sequence, captured at 100~FPS. The top graph shows the full depth map (obtained in histogram mode), the bottom graph illustrates selective readout according to changes in time gate position. See also Visualisation \href{https://youtu.be/BCli817EKk0}{3}.
}
\label{fig:pass}
\end{figure}

Figs.~\ref{fig:swing} and \ref{fig:pass} present snapshots of data from dynamic scenes. In Fig.~\ref{fig:swing}, a person is walking around a moving swing. Two depth maps are shown: sub-bin resolution depth, extracted off-chip from histogram data, and bin-resolution depth, as reported by the sensor. Even though the latter output is one tenth of the size compared to histogram frames, and provides a very coarse depth reading given the large bin size of 8~ns, it is still seen to capture key features in the scene. Fig.~\ref{fig:pass} gives a demonstration of the smart readout mode, where only pixels with changing depth are reported, and the rest replaced by zeros. On the top, we see the original sequence, of two people passing a ball to each other, and on the bottom we see the result of only presenting pixel data for pixels with a change in time gate position. This gives a “dynamic vision sensor”-type output \cite{delbruck2010}, with the movement of the ball clearly indicated.

Videos of these examples, and of another garden scene where the camera is panned, can be found in the supplementary material (Visualisations \href{https://youtu.be/qlIr0gX-t2M}{1}, \href{https://youtu.be/6YGAO-eOoWM}{2}, \href{https://youtu.be/BCli817EKk0}{3} and
\href{https://youtu.be/fG32Wp22QCE}{4}). In addition, two longer range sequences are presented. In the first example, a person is walking towards the camera in an open space, with pixels observing the person successfully tracking the motion in z (Visualisations \href{https://youtu.be/S1anZX_BxOw}{5} and \href{https://youtu.be/YKYxvZ6kW0g}{6}). The second example (Visualisation \href{https://youtu.be/uhOiCV2-MGE}{7}) compares sequences captured in the tracking and sliding (continuous scanning) modalities using the same exposure time. In the sliding mode, the camera steps through time gate positions 1 to 16, and the first peak detected by each pixel is plotted. Both sequences are played back in real time. In the tracking mode, a $\times 16$ higher frame rate is achieved, which could potentially be exploited in post-processing to obtain more accurate depth or motion estimates in z for tracked surfaces. No post-processing has been applied to any of the data, apart from centre-of-mass peak extraction (where indicated), and zero-offset correction, based on subtracting an a priori depth image of a flat calibration target.

\subsection*{Power consumption}

The chip power consumption in outdoor operation at 50~FPS under 30~klux of ambient light was measured to be 70~mW, of which 32~mW is consumed by SPAD detectors, 36~mW by the digital processing logic, and 1.32~mW by the I/O pads. The latter, I/O power consumption is strongly dependent on the frame rate, as well as the readout mode. In the default histogram mode, power consumption ranges from 1.32~mW-11.6mW (50~FPS-1~kFPS), reducing to 0.5~mW-1.85~mW in CMM mode. Further reductions can be obtained using the smart readout modes, which can lower the I/O power consumption to 0.43~mW-1.62~mW in histogram mode and 0.26~mW-0.99~mW in CMM mode. These results demonstrate the effectiveness of the reduced readout modes in saving I/O power, with additional savings expected from the row skipping feature. The chip was fabricated in STMicroelectronics' general purpose 40~nm CMOS technology; a SPAD detector power improvement may be made in future versions of this chip by using STMicroelectronics' SPAD-optimised technology which would lead to lower DCR and <100 fC per SPAD pulse \cite{pellegrini2022}.

\section{Discussion and Outlook}

A SPAD dToF imager has been presented that the authors believe to be the first reported imager with in-pixel ambient estimation, surface detection and tracking, resulting in a scalable architecture.

Each pixel has an independent time gate that automatically tracks the signal peak, and histogramming logic to generate photon timing histograms. Multiple events can be captured per laser cycle to enable outdoor operation without pile-up distortion. Thanks to the significant in-pixel data compression, the device can operate at high-frame rates, and on-chip depth computation is available. Experimental results are given demonstrating the viability of the device for medium range LIDAR at high frame rates.

Although not considered here, there is the potential to enhance the time gate mechanism in each pixel by switching to external control of the time gate positions, for instance, with the intention to capture multiple peaks. Such external control would also allow for more accurate background estimation and peak detection, for example, by capturing pairs of exposures, one with the laser off and the other with the laser on.

There is also scope for more optimised illumination strategies compared with the simple flood illumination at fixed repetition rate that is currently adopted. As described in \cite{taneski2022}, when stepping through different time gate positions, it is beneficial to adjust the laser power accordingly. In particular, a higher laser repetition rate (or higher pulse energy) is advocated for farther time window positions (to ensure a sufficient SNR for the histogram peaks from far away targets to be reliably detected), and lower repetition rate (or lower pulse energy) for closer time window positions to conserve laser power. In the context of the present sensor where each pixel has its own time gate position, this would require an illumination source with individually addressable elements, allowing the number of laser cycles to be adjusted on a per-pixel basis.

State-of-the-art SPADs in 3D-stacked, backside illluminated (BSI) technology \cite{ito2020,morimoto2021,pellegrini2022} offer significantly higher photon detection efficiency (PDP) compared with the front-side illuminated (FSI) technology used in the present chip. By combining the latest SPADs with the processing architecture presented here, the same signal (and background) counts, and therefore depth precision, could be attained at much lower exposure times, and consequently, at higher frame rates. 

\mbox{}
\bibliographystyle{ieeetr}
\bibliography{HSLIDAR}

\section*{Funding}
The support of the Engineering and Physical Science Research Council (EPSRC) via grants EP/M01326X/1 and EP/S001638/1 is acknowledged.

\section*{Acknowledgments}
The authors are grateful to STMicroelectronics for chip fabrication and in particular to Sara Pellegrini and Thierry Lachaud for their support.  Portions of this work were presented at the International Image Sensor Workshop in 2021 \cite{gyongy2021}.

\section*{Disclosures}
The authors declare no conflicts of interest.

\section*{Data availability} 
Data underlying the results presented in this paper are not publicly available at this time but may be obtained from the authors upon reasonable request.

\end{document}